\documentclass[ 12pt, numbers,sort&compress]{elsarticle}

\usepackage{amssymb}
\usepackage{graphicx}
\usepackage{subfigure}
\usepackage{float}
\usepackage{amsmath}
\usepackage{booktabs}
\usepackage{multirow}
\usepackage{makecell,rotating}
\usepackage{url}

\begin{document}

\begin{frontmatter}



\title{A Graph-Based Semi-Supervised $k$ Nearest-Neighbor Method for Nonlinear Manifold Distributed Data Classification}


\author[RRLab]{Enmei Tu\corref{cor1}}
\author[NTU]{Yaqian Zhang}
\author[ZL]{Lin Zhu}
\author[SJTU]{Jie Yang}
\cortext[cor1]{Corresponding author: Enmei Tu, hellotem@hotmail.com}
\author[AUT]{Nikola Kasabov}
\address[RRLab]{Rolls-Royce@NTU Corporate Lab, Nanyang Technological University, Singapore}
\address[NTU]{School of Computing Engineering, Nanyang Technological University, Singapore}
\address[ZL]{School of Computer Science and Technology, Shanghai University of Electric Power, China}
\address[SJTU]{Institute of Image Processing and Pattern Recognition, Shanghai Jiao Tong University, China}
\address[AUT]{The Knowledge Engineering and Discovery Research Institute, Auckland University of Technology, New Zealand}

\begin{abstract}
$k$ Nearest Neighbors ($k$NN) is one of the most widely used supervised learning algorithms to classify Gaussian distributed data, but it does not achieve good results when it is applied to nonlinear manifold distributed data, especially when a very limited amount of labeled samples are available. In this paper, we propose a new graph-based $k$NN algorithm which can effectively handle both Gaussian distributed data and nonlinear manifold distributed data. To achieve this goal, we first propose a constrained Tired Random Walk (TRW) by constructing an $R$-level nearest-neighbor strengthened tree over the graph, and then compute a TRW matrix for  similarity measurement purposes. After this, the nearest neighbors are identified according to the TRW matrix and the class label of a query point is determined by the sum of all the TRW weights of its nearest neighbors. To deal with online situations, we also propose a new algorithm to handle sequential samples based a local neighborhood reconstruction. Comparison experiments are conducted on both synthetic data sets and real-world data sets to demonstrate the validity of the proposed new $k$NN algorithm and its improvements to other version of $k$NN algorithms. Given the widespread appearance of manifold structures in real-world problems and the popularity of the traditional $k$NN algorithm, the proposed manifold version $k$NN shows promising potential for classifying manifold-distributed data.
\end{abstract}


\end{frontmatter}



\section{Introduction}
\begin{figure}[H]
  \centering
   \subfigure[Toy manifold data set]{
   \includegraphics[width=0.4\textwidth]{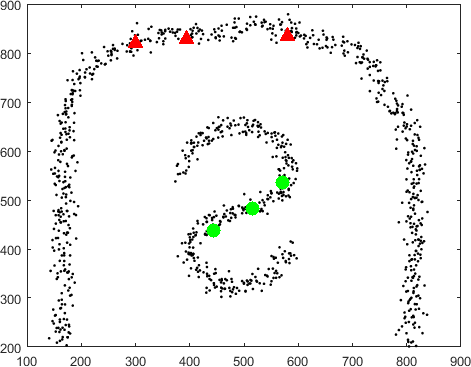}}
   \subfigure[$k$=1]{
   \includegraphics[width=0.4\textwidth]{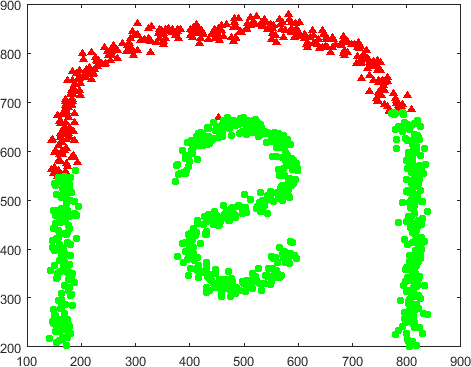}}

   \subfigure[$k$=2]{
   \includegraphics[width=0.4\textwidth]{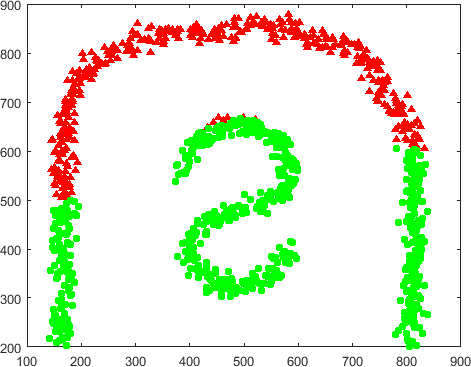}}
  \subfigure[$k$=3]{
  \includegraphics[width=0.4\textwidth]{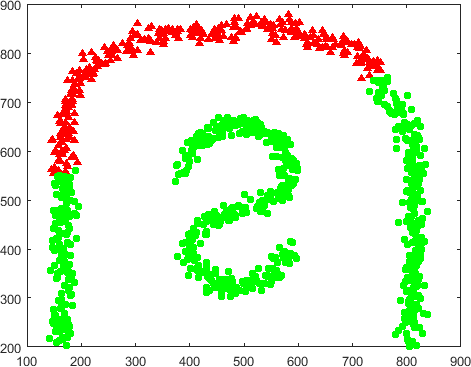}}
  \caption{Results of traditional $k$NN classification with $k$=1, 2, 3 on manifold distributed data, in which the colored shapes (red triangles and green dots) are labeled samples and the black dots are unlabeled samples.}
  \label{USData}
\end{figure}

$k$ Nearest Neighbors ($k$NN) \cite{cover1967nearest, wu2008top, samanthula2015k, xie2016robust} is one of the most popular classification algorithms and has been widely used in many fields, such as intrusion detection \cite{liao2002use}, gene classification \cite{li2001gene}, semiconductor fault detection \cite{he2007fault}, very large database manipulation \cite{kolahdouzan2004voronoi},  nuclear magnetic resonance spectral interpretation \cite{kowalski1972k} and the prediction of basal area diameter \cite{maltamo1998methods}, because it is simple but effective, and can generally obtain good results in many tasks. One main drawback of the traditional $k$NN is that it does not take the manifold distribution information into account and this can cause bias which results in bad performance. It becomes even worse when there are only a very small amount of labeled samples available. To address this, an example is shown in figure \ref{USData}(a), in which there are two one-dimensional manifolds (the outer arch and the interior reflected \emph{S}) which correspond to two classes, respectively. Each class has only 3 labeled samples, indicated by the colored triangles and circles. Black dots are unlabeled samples. Figure \ref{USData}(b) - (d) show the 1NN, 2NN and 3NN classification results produced by the traditional $k$NN, respectively. We can see that although the data have apparent manifold distribution, the traditional $k$NN incorrectly classifies many samples due to ignoring the manifold information.

To improve the performance of the traditional $k$NN, some new $k$NN algorithms have been proposed. Hastie et. al. \cite{hastie1996discriminant} proposed an adaptive $k$NN algorithm which computes a local metric for each sample and uses Mahalanobis distance to find the nearest neighbors of a query point. Hechenbichler and Schliep \cite{hechenbichler2004weighted} introduced a weight scheme to attach different importance to the nearest neighbors with respect to their distances to the query point. To reduce the effect of unbalanced training set sizes of different classes, Tan \cite{tan2005neighbor} used  different weights for different classes regarding  the number of labeled samples in each class. There are also some other improvements and weight schemes for different tasks \cite{li2003improved,cost1993weighted,nene1997simple,weinberger2009distance,dudani1976distance,han2001text,gao2016reverse}.

However, none of these new $k$NN algorithms takes manifold structure into consideration explicitly. For high-dimensional data, such as face images, documents and video sequences, the nearest neighbors of a point found by traditional $k$NN algorithms can be very far in terms of  the geodesic distance between them, because the dimension of the underlying manifold is usually much lower than  that of the data space \cite{roweis2000nonlinear,tenenbaum2000global, beyer1999nearest}.  There have also been attempts to make $k$NN adaptive to manifold data. In Turaga and Chellappa's paper \cite{turaga2010nearest}, geodesic distance is used to directly replace standard Euclidean distance in traditional $k$NN, but geodesic distance can be computed with good accuracy only if the manifold is sampled with sufficient points. Furthermore, geodesic distance  tends to be very sensitive to short-circuit phenomenon. Li \cite{ma2010local} proposed a weighted manifold $k$NN using Local Linear Embedding (LLE) techniques, but LLE tends to be unstable due to local changes on the manifold. Percus and Olivier \cite{percus1998scaling} studied the general $k^\text{th}$ nearest neighbor distance metric on close manifold, but their method needs to know exactly the analytical form of the manifold and thus is unsuitable for most real-world applications.

In this paper, we propose a novel graph-based $k$NN algorithm which can effectively handle both traditional Gaussian distributed data and nonlinear manifold distributed data. To do so, we first present a manifold similarity measure method, the constrained tired random walk, and then we modify the traditional $k$NN algorithm to adopt the new measuring method. To deal with online situations, we also propose a new algorithm to handle sequential samples based on a local neighborhood reconstruction method. Experimental results on both synthetic and real-world data sets are presented to demonstrate the validity of the proposed method.

The remainder of this paper is organized as follows: Section 2 reviews the tired random walk model. Section 3  presents a new constrained tired walk random walk model and Section 4 describes the graph-based $k$NN algorithm. Section 5 proposes a sequential algorithm for online samples. The simulation and comparison results are presented in Section 6, followed by conclusions in Section 7.
\section{ Review of tired random walk}
Assume a  training set ${{\mathcal{X}}_{T}}=\left\{ {{x}_{1}},{{x}_{2}},...,{{x}_{l-1}},{{x}_{l}} \right\}\subset {{\mathbb{R}}^{d}}$  contains $l$ labeled samples and the class label of ${{x}_{i}}$ is ${{y}_{i}}$ , ${{y}_{i}}\in \left\{ 1,2,...,C \right\}; i=1...l$, where $d$ is feature length and $C$ is class number. There are also $n-l$ unlabeled samples to be classified, ${{\mathcal{X}}_{U}}=\left\{ {{x}_{1}},{{x}_{2}},...,{{x}_{n-l}} \right\}$.  Denote $\mathcal{X}={{\mathcal{X}}_{T}}\cup {{\mathcal{X}}_{U}}=\left\{ {{x}_{1}},{{x}_{2}},...,{{x}_{n}} \right\}$ and $y=(y_1,y_2,...,y_l)$. We also use $X$ for the sample matrix, whose columns are the samples in $\mathcal{X}$.  We use $\left\| \cdot  \right\|$ to denote  the Frobenius norm.

In differential geometry studies, a manifold can be defined from an intrinsic or extrinsic point of view \cite{spivak1970comprehensive, boothby2003introduction}. But in data processing studies, such as dimension reduction \cite{chahooki2014shape, petraglia2015dimensional, nowakowska2016dimensionality} and manifold learning \cite{vincent2002manifold, bengio2005non, vemulapalli2013kernel}, it is helpful to consider a manifold as a distribution which is embedded in a higher Euclidean space, i.e. adopting an extrinsic view in its ambient space. Borrowing the concept of \emph{intrinsic dimension} from \cite{camastra2016intrinsic},  we  give a formal definition of a manifold data set as follows.\\
\emph{\textbf{Definition}: A data set is considered to be manifold distributed if its intrinsic dimension is less than its data space dimension.}\\
For more information on the intrinsic dimensions of a data set and how it can be estimated from data samples, we refer readers to  \cite{camastra2016intrinsic, petraglia2015dimensional}. To determine whether a data set has manifold distribution, one can simply estimate its intrinsic dimension and compare it with the data space dimension, i.e. the length of a sample vector.

Similarity measure is an important factor while processing manifold distributed data, because on manifolds traditional distance metrics  (such as Euclidean distance) are not a proper measure \cite{yu2014semantic,  yi2016online, yu2016density}. Recent studies have also proved that classical random walk is not a useful measure for large sample  cases or high-dimensional data because it does not take any global properties of the data into account \cite{luxburg2010getting}. The tired random walk (TRW) model was proposed in Tu's paper \cite{tu2014novel} and has been demonstrated to be an effective measure of nonlinear manifold \cite{wang2015characterizing,yin2015self}, because it takes global geometrical structure information into consideration.

\begin{figure}[H]
  \centering
   \includegraphics[width=0.5\textwidth]{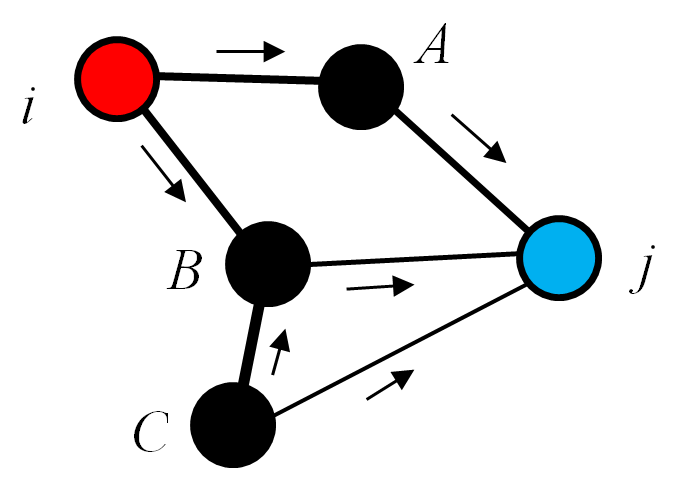}
  \caption{Tired random walk on weighted undirected graph.}
  \label{TRWGraph}
\end{figure}

Recall that  on a  weighted undirected graph,  the classical random walk transition matrix is $P={{D}^{-1}}W$, where $W$ is the graph adjacency matrix and $D$ is a diagonal matrix with entries ${{D}_{ii}}=\sum\nolimits_{j=1}^{n}{{{W}_{ij}}}$.  Now imagine that a tired random walker walks continuously through edges in a graph, but it becomes more tired after each walk and finally stops after all energy is exhausted, i.e. the transition probability of the random walk reduces with a fixed ratio (e.g. 0.01) after each walk and finally approaches 0. After $t$ steps the tired random walk transition probability matrix becomes ${{(0.01P)}^{t}}$.  Now considering figure \ref{TRWGraph}, the tired random walker starts from vertex ${i}$ and its destination is vertex ${j}$ on the graph, walking with a strength reduction rate $\alpha \in (0,1)$. Then it may walk through any path that connects vertices ${i}$ and ${j}$, with an arbitrary number of steps before its strength is used up. For example, the tired random walker can walk through path $i \rightarrow A\rightarrow j$, or $i \rightarrow B\rightarrow j$. It can also walk through $i \rightarrow B \rightarrow C  \rightarrow j$,  or even $i \rightarrow B \rightarrow C  \rightarrow B \rightarrow j$, because while at vertex $C$, the probability of walking to $B$ is much larger than that to $j$.  But all these walking paths reflect the underlying geometrical structure of the graph, hence the distribution of the data. Therefore,  a good similarity measure between vertex $i$ to $j$  should  take (globally) all  possible paths and an arbitrary number of  steps  (can potentially be infinite) into consideration, rather than only considering (locally) a single path or a single step as the classical random walk \cite{luxburg2010getting} does. This makes a fundamental difference between the \emph{tired random walk} and \emph{classical random walk} and entitles the tired random walk to be more robust and effective, especially for manifold distributed data, as will be demonstrated in the experimental section.  Mathematically, the accumulated transition probability of the tired random walk between vertex $i$ and $j$ is ${{{(P_{TRW})}}_{ij}}={{\left[ \sum\nolimits_{t=0}^{\infty }{{{(\alpha P)}^{t}}} \right]}_{ij}}$. For all vertices, the accumulated transition probability matrix becomes ${P_{TRW}}=\sum\nolimits_{t=0}^{\infty }{{{(\alpha P)}^{t}}}$. As the eigenvalue of $P$ is in $[-1,\ 1]$ and $\alpha \in (0,1)$, the series converges and the tired random walk matrix is
\begin{equation} \label{PTRWEquation}
{{P}_{TRW}}=\sum\nolimits_{t=0}^{\infty }{{{\left( \alpha P \right)}^{t}}}={{\left( I-\alpha P \right)}^{-1}}
\end{equation}

 Because ${{P}_{TRW}}$ takes all the possible paths into account, it captures the global geometrical structure information of the underlying manifold and thus is demonstrated to be more effective and robust to describe manifold similarity.
\section{A constrained tired random walk model}

In this section, we further extend the model into a constrained situation. For classification purposes, we find that labeled samples provide not only  class distribution information, but also constraint information, i.e., samples which have the same class labels are must-link pairs and samples which have different class labels are cannot-link pairs. In most of the existing supervised learning algorithms, only  class information is utilized but  constraint information is discarded. Here we include  constraint information into the TRW model by modifying the weights of graph edges between the labeled samples, because constraint information has been demonstrated to be useful for performance improvement \cite{tu2015posterior, gong2014semi, fu2015local, zhou2015semi}. Class information will be utilized in the next section for the proposed new $k$NN algorithm.

We first construct an $R$-level nearest-neighbor strengthened tree for each labeled sample ${{x}_{i}}\in {{\mathcal{X}}_{T}}$ as follows:
\begin{enumerate}[(i).]
  \item Set ${{x}_{i}}$ as the first level node (tree root node, $r=0$) and its $k$ nearest neighbors as the second level  nodes.
  \item For each node in level $r-1$, set its nearest neighbors  as its level $r$ descendants. If any node in level $r$  appears in its ancestor level, remove it from  level $r$.
  \item	If $r<R$, go to (ii).
\end{enumerate}
where $R$ is a user-specified parameter to define the depth of the tree.  Then for each pair of samples $\left( {{x}_{i}},{{x}_{j}} \right)$, the corresponding graph edge weight is set according to the rules in Table \ref{ConstrainedGraph}.
 \begin{table}[ht]
\caption{Constrained graph construction}
\label{ConstrainedGraph}
\centering  
\footnotesize
\begin{tabular}{ll}  
\toprule \hline
Steps & \\ \hline  
1 &	${{W}_{ij}}=1$ if ${{x}_{i}},{{x}_{j}}\in {{\mathcal{X}}_{T}}$  and have same class label.\\
2 & ${{W}_{ij}}=0$ if ${{x}_{i}},{{x}_{j}}\in {{\mathcal{X}}_{T}}$ and have different class labels. \\
3 &${{W}_{ij}}=\exp \left( -{{{\left\| {{x}_{i}}-{{x}_{j}} \right\|}^{2}}}/{2{{\sigma }^{2}}}\; \right)$ if at least one of ${{x}_{i}}$, ${{x}_{j}}$ is unlabeled. \\
4 &${{W}_{ij}}=(1+\theta _{ij}^{r}){{W}_{ij}},1\le r\le R$, if ${{x}_{i}}$ is a node in level $r-1$  and ${{x}_{j}}$ \\
 &is a child of $x_i$ in level $r$ in the strengthened tree. \\ \hline \bottomrule        
\end{tabular}
\end{table}
$\sigma $ is the Gaussian kernel width parameter and $\theta$ is the strengthening parameter. Note that for the Gaussian kernel, step 1 is equivalent to merging the two same-label samples together (0 distance between them) and step 2 is equivalent to separating  the two different-label samples from each other to infinitely far (infinite distance between them). The connections from a labeled sample to its nearest neighbors are strengthened by ${{\theta }_{ij}}$ in step 4. This can spread the ‘hard’ constraints in steps 1 and  2 to farther neighborhoods on the graph in a form of \emph{soft} constraints and thus causes these constraints to have a wider influence. The motivation of constructing the strengthened tree is inspired by the neural network reservoir structure analysis techniques, in which information has been shown to spread out from input neurons to interior neurons in the reservoir following a tree-structure path \cite{tu2016mapping}.

The selection of parameter $\theta$ is based the following conditions
\begin{itemize}
\item the strengthened weight  should be positive and less than the weight of the must-link constraint.
\item  the strengthening effect should be positive and decays along the strengthened tree level.
\end{itemize}
Mathematically, the conditions are
\begin{equation}
\left\{ \begin{aligned}
  & 0<{{W}_{ij}}+{{\theta }_{ij}}{{W}_{ij}}<1 \\
 & 0<\theta _{ij}^{r+1}<\theta _{ij}^{r} \\
\end{aligned} \right.
\end{equation}
As a result, $\theta$ should be
\begin{equation}
0<{{\theta }_{ij}}<\min \left( \frac{1-{{W}_{ij}}}{{{W}_{ij}}},1 \right)=\bar{\theta }
\end{equation}
In all our experiments, we used a single value of $\theta=0.1\bar{\theta }$, which
gives good results for both synthetic and real-world data.
\section{A new graph-based $k$NN classification algorithm on nonlinear manifold}

Here we present a graph-based $k$NN algorithm for nonlinear manifold data  classification. The procedure of the algorithm is summarized in Table \ref{GraphKNN}.
 \begin{table}[ht]
\caption{A graph-based $k$NN algorithm}
\label{GraphKNN}
\centering  
\footnotesize
\begin{tabular}{ll}  
\toprule \hline
Steps & \\ \hline  
1 &Input $\mathcal{X}={{\mathcal{X}}_{T}}\cup {{\mathcal{X}}_{U}}$, $y$ and $k$. \\
2 &Construct a constrained graph according to Table \ref{ConstrainedGraph} \\
3 &Compute the ${P}_{TRW}$ using equation (\ref{PTRWEquation}) \\
4 &Evaluate samples' similarity according to equation (\ref{PTRW}) \\
5 &Find nearest neighbors of an unlabeled sample using equation (\ref{FindNN}) \\
6 &Determine the class label according to equation (\ref{KNNClassification}) \\ \hline  \bottomrule      
\end{tabular}
\end{table}

Specifically, given ${{P}_{TRW}}$ matrix, the TRW weight between sample ${{x}_{i}}$ and ${{x}_{j}}$ is defined as
\begin{equation} \label{PTRW}
\bar w_{ij}=w({{x}_{i}},{{x}_{j}})=\frac{{{({{P}_{TRW}})}_{ij}}+{{({{P}_{TRW}})}_{ji}}}{2}
\end{equation}
Note that while the similarity measure defined by matrix $P_{TRW}$ between two samples is not necessarily symmetric ( $P$ is not symmetric and thus its matrix series $P_{TRW}$ is also not symmetric), the weight defined in equation (\ref{PTRW}) is indeed a symmetric measure. For each unlabeled sample $x\in {{\mathcal{X}}_{U}}$, we could find its $k\le l$ nearest  neighbors  from ${{\mathcal{X}}_{T}}$ by
\begin{equation}\label{FindNN}
{{x}_{i}}=\underset{{{x}_{j}}\in {{\mathcal{X}}_{T}}}{\mathop{\arg \max }}\,w(x,{{x}_{j}})
\end{equation}
Instead of counting the number of labeled samples from each class in the classical $k$NN, we sum  the TRW weights of the labeled samples of each class and the class label of the unlabeled sample $x$ is determined by
\begin{equation}\label{KNNClassification}
y=\underset{c=1,2,...,C}{\mathop{\arg \max }}\,\sum\nolimits_{i=1}^{k}{w(x,{{x}_{i}})I({{y}_{i}}=c)}
\end{equation}

It is worth mentioning that because the proposed semi-supervised m$k$NN  utilizes class label information only in the classifying stage and it uses the same $k$ value equally for all classes, m$k$NN naturally lacks  the so-called class bias problem in many semi-supervised algorithms (such as \cite{zhu2003semi, liu2010large,tu2013experimental,  ji2014automatic}) that is due to the influence of unbalanced labeled samples of each class in $\mathcal{X}_T$ and needs to be re-balanced by various weighting schemes \cite{wang2013semi, zhu2003semi}. Furthermore, with the algorithm described in the next section, m$k$NN immediately becomes  a supervised  classifier and enjoys the advantages of both semi-supervised learning (classifying abundant unlabeled samples with only a tiny number of labeled samples) and supervised learning (classifying new samples immediately without repeating the whole learning process).

\section{A sequential method to handle online  samples}
For one single unlabeled sample, if we compute its TRW weights using equation (\ref{PTRWEquation}) and (\ref{PTRW}), we have to add it to $\mathcal{X}$ and recompute the matrix ${{P}_{TRW}}$. As a result, the computational cost for \emph{one} sample is too high. Actually this is the so-called transductive learning problem\footnote{Transductive learning is an opposite concept to inductive learning. Inductive learning means the learning algorithm, such as SVM, learns a model explicitly in  data space that partitions the data space into several different regions. Then the model  can be applied directly to unseen samples to obtain the class labels. On the other side, transductive learning does not build any model. It performs one-time learning only on a fixed data set.Whenever the data set changes (for example, existing samples are changed or new samples are added.), the whole learning process has to be repeated again to assign new class labels.},  a common drawback of many existing algorithms \cite{joachims1999transductive,joachims2003transductive,collobert2006large,goldberg2010transduction}. To attack this problem, we propose a new method based on rapid neighborhood reconstruction, in which a local neighborhood is first constructed in sample space and then the TRW weights can be reconstructed in the same local neighborhood with very trivial computational cost.

Given a new sample $x$, it has been shown that $x$ can be well reconstructed by its nearest neighbors on the manifold if there are sufficient data points sampled from the manifold \cite{roweis2000nonlinear, chen2013nonnegative, tao2015local}. Thus, it is also reasonable to assume that the neighborhood relationships, hence the weights of sample $x$  in equation (\ref{PTRW}), have the same geometrical distribution as the sample distribution. So, to compute the weights of $x$ without explicitly recomputing matrix $P_{TRW}$ in equation (\ref{PTRW}), we first find $x$'s $k$ nearest neighbors\footnote{One should note that finding nearest neighbors in $\mathcal{X}$ is quite different from that in $\mathcal{X}_T$. The former is the basis of many nearest-neighbor operations (such as constructing nearest-neighbor graph in \cite{tenenbaum2000global, roweis2000nonlinear} and the $R$-level strengthened tree in Section 3 of this paper) and the latter is the basis of the classical $k$NN classifier. Because $\mathcal{X}$ contains many instances, which are  sampled densely from the underlying data distribution, an instances's local neighborhood in $\mathcal{X}$ is usually very small  and thus Euclidean distance is still valid in this small range for that any manifold can be locally well approximated by Euclidean space \cite{boothby2003introduction}.  However, instances in $\mathcal{X}_T$ are very few and usually not densely sampled from the manifold and \emph{nearest neighbors }in $\mathcal{X}_T$ can be very far. Thus Euclidean distance is no longer suitable for measuring the closeness of the points in $\mathcal{X}_T$. This is why we need other new similarity (or closeness) measure methods, which is one of the main contributions of this paper.} in $\mathcal{X}$, written as $X_{k}$ which contains these $k$ nearest neighbors in its columns, and then minimize the local reconstruction error by solving the following constraint quadratic optimization problem
\begin{equation}
\begin{aligned}\label{OptWeight}
  & \min {{\left\| x-X_{k}z \right\|}^{2}} \\
 & s.t.\,\,\,z\ge 0;{{z}^{T}}e=1 \\
\end{aligned}
\end{equation}
where $e$ is a vector with all entries being 1. Note that the entries of $z$ are nonnegative and the sum of all entries must be 1, so $z$ is expected to be sparse \cite{liu2010large}. Problem (\ref{OptWeight}) is a constrained quadratic optimization problem and can be solved very efficiently with many publicly available toolboxes. We use the OPTI optimization toolbox\footnote{OPTI TOOLBOX: http://www.i2c2.aut.ac.nz/Wiki/OPTI/} to solve this. After obtaining the optimal $z$, the TRW weights between $x$ and its nearest labeled samples in ${{X}_{k}}$ can be computed by solving the following optimization problem
\begin{equation}\label{ReconstructWeight}
\begin{aligned}
  & \min {{\left\| \bar w-\bar W_{k}^{T}z \right\|}^{2}} \\
 & s.t.\quad \bar w\ge 0 \\
\end{aligned}
\end{equation}
where ${{\bar W}_{k}}=(\bar w_1, \bar w_2,...,\bar w_k)$ and $\bar w_i$ contains  the weights between sample $x_i$ in $X_{k}$ and its $k$ nearest neighbors in $X$.

\begin{figure}[H]
  \centering
   \subfigure[Computing reconstruction weight $z$ using equation (\ref{OptWeight})]{
   \includegraphics[width=0.45\textwidth]{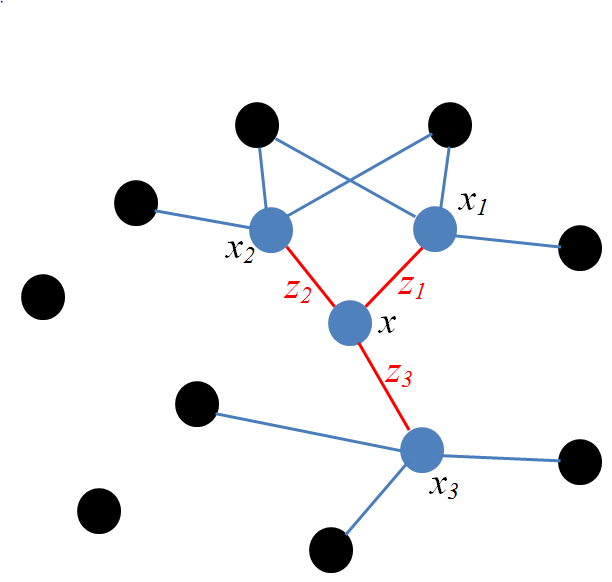}}
 \subfigure[Computing TRW weight $\bar w$ using equation (\ref{ReconstructWeight})]{
   \includegraphics[width=0.45\textwidth]{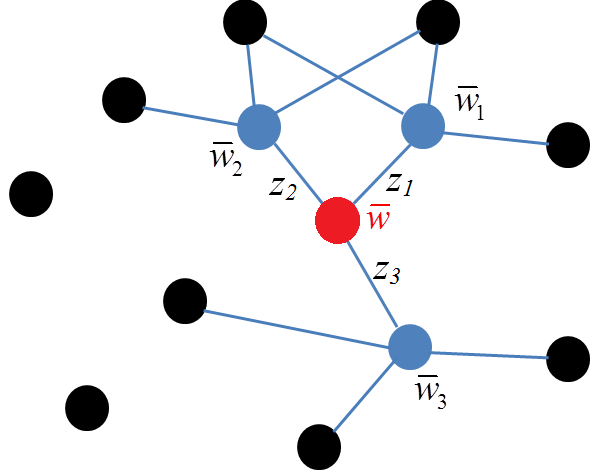}}
  \caption{The process of computing sequential TRW weight.}
  \label{IllustrationExample}
\end{figure}

This process is illustrated in Figure \ref{IllustrationExample}. Figure \ref{IllustrationExample}(a) corresponds to equation (\ref{OptWeight}) which computes $z$ and Figure \ref{IllustrationExample}(b) corresponds to equation (\ref{ReconstructWeight}) which computes $\bar w$.

It is easy to see that the optimal solution of problem (\ref{ReconstructWeight}) is simply the result of a nonnegative projection operation
\begin{equation}
\bar w=\text{max}(0, \bar W_{k}^{T}z)= \bar W_{k}^{T}z
\end{equation}
The second equation holds because both $\bar W_{k}$ and $z$ are nonnegative and therefore their multiplication result is also nonnegative. 


One should note that with this sequential learning strategy, the proposed m$k$NN can be treated as an inductive classifier, whose model consists of both the $P_{TRW}$ matrix and the data samples that have been classified so far. Whenever a new sample arrives, the model can quickly give its class label by the three  steps in Table \ref{SKNN}, without repeating the whole learning process in Table \ref{GraphKNN}.

 \begin{table}[ht]
\caption{The procedure of sequential manifold $k$NN algorithm}
\label{SKNN}
\centering  
\footnotesize
\begin{tabular}{ll}  
\toprule \hline
Steps & \\ \hline  
1 &Input $\mathcal{X}={{\mathcal{X}}_{T}}\cup {{\mathcal{X}}_{U}}$, $P_{TRW}$, $x$ and $k$. \\
2 &Find $x$'s  $k$ nearest neighbors in $\mathcal{X}$  \\
3 &Use equations (\ref{OptWeight}) and  (\ref{ReconstructWeight}) to compute its TRW weights \\
4 &Use  equation (\ref{KNNClassification}) to classify it \\ \hline  \bottomrule      
\end{tabular}
\end{table}

\section{Experimental results}
In this section, we report the experimental results on both the synthetic data sets and real-world data sets. The comparison algorithms include traditional $k$ nearest neighbors ($k$NN), the weighted $k$ nearest neighbors (w$k$NN) and the geodesic $k$NN (g$k$NN) proposed by Pavan and Rama\cite{turaga2010nearest}, as well as  our manifold $k$ nearest neighbors (m$k$NN).  For $k$NN and w$k$NN, the only parameter is $k$. For g$k$NN and m$k$NN, there is one more parameter for each, i.e. the number of nearest neighbors for computing geodesic distance in g$k$NN and the kernel width $\sigma $ in m$k$NN. We tune these two parameters by grid-search and choose their values to produce the minimal 2-fold cross validation error rate.
\subsection{	Experimental results on synthetic data sets}
\begin{figure}[H]
  \centering
   \subfigure[data set 1]{
   \includegraphics[width=0.3\textwidth]{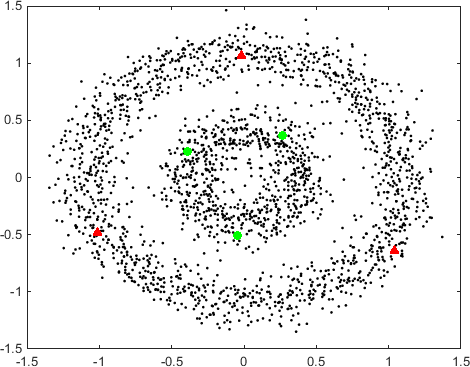}}
   \subfigure[data set 2]{
   \includegraphics[width=0.3\textwidth]{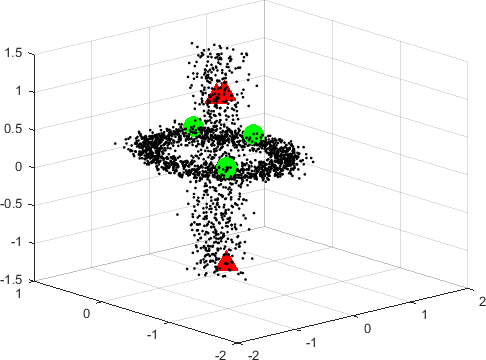}}
   \subfigure[data set 3]{
   \includegraphics[width=0.3\textwidth]{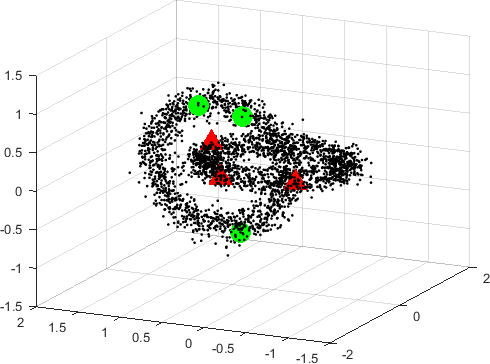}}
  \caption{Three synthetic data sets, in which the colored shapes are labeled samples and the black dots are unlabeled samples}
  \label{SynDataSet}
\end{figure}

\begin{figure}[H]
  \centering
  \subfigure[$k$NN]{
   \includegraphics[width=0.4\textwidth]{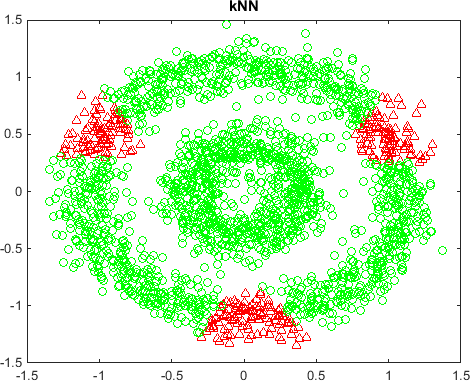}}
   \subfigure[w$k$NN]{
   \includegraphics[width=0.4\textwidth]{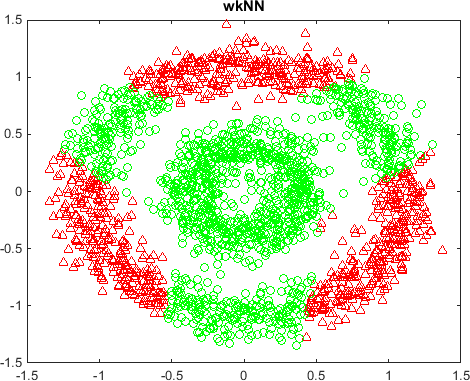}}

    \subfigure[g$k$NN]{
   \includegraphics[width=0.4\textwidth]{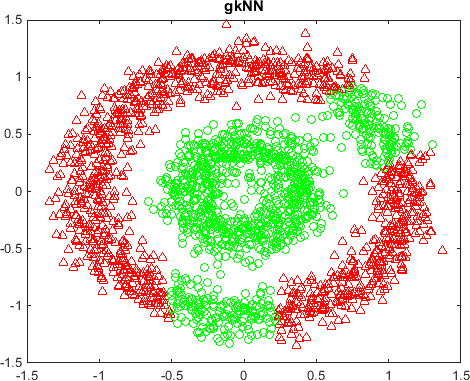}}
    \subfigure[m$k$NN]{
   \includegraphics[width=0.4\textwidth]{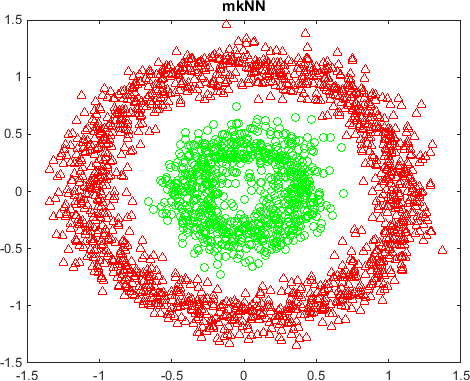}}
  \caption{Experimental results on synthetic data sets. (a)-(d): results of $k$NN, w$k$NN, g$k$NN and m$k$NN on data set 1.}
  \label{ResultTwoCircle}
\end{figure}

\begin{figure}[H]
  \centering
   \subfigure[$k$NN]{
   \includegraphics[width=0.43\textwidth]{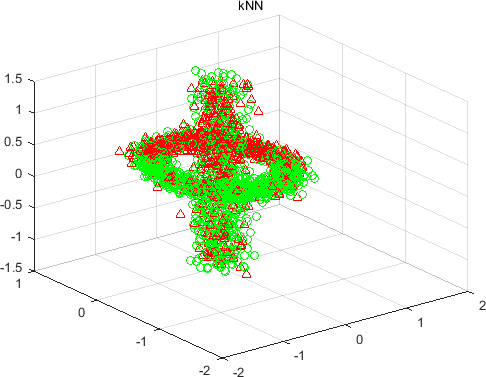}}
   \subfigure[w$k$NN]{
   \includegraphics[width=0.43\textwidth]{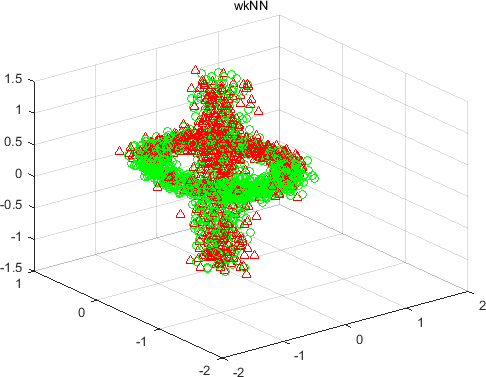}}

    \subfigure[g$k$NN]{
   \includegraphics[width=0.43\textwidth]{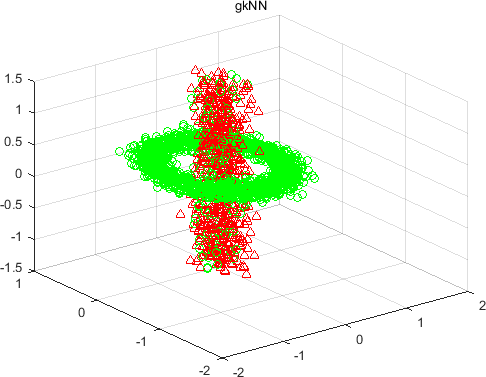}}
    \subfigure[m$k$NN]{
   \includegraphics[width=0.43\textwidth]{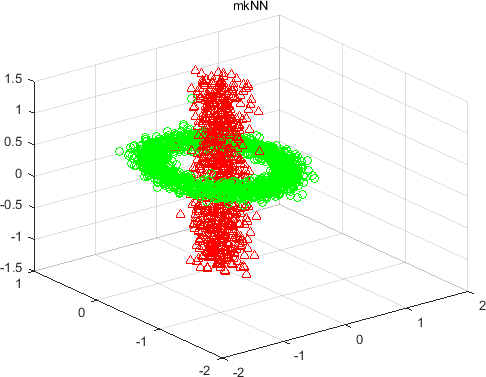}}
    \caption{Experimental results on synthetic data sets. (a)-(d): results of $k$NN, w$k$NN, g$k$NN and m$k$NN on data set 2.}
    \label{Resultcirclerub}
\end{figure}

\begin{figure}[H]
  \centering
   \subfigure[$k$NN]{
   \includegraphics[width=0.43\textwidth]{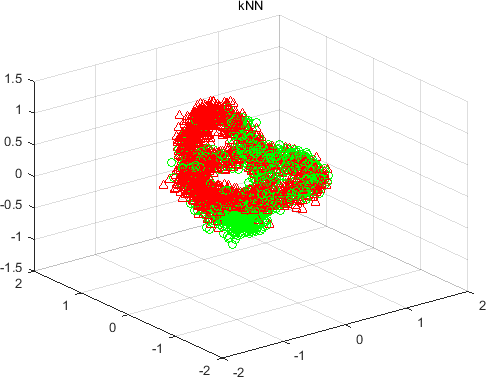}}
   \subfigure[w$k$NN]{
   \includegraphics[width=0.43\textwidth]{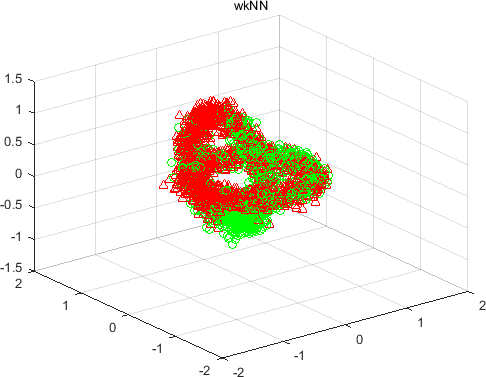}}

    \subfigure[g$k$NN]{
   \includegraphics[width=0.43\textwidth]{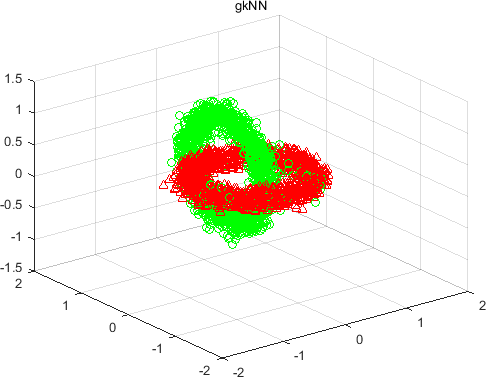}}
    \subfigure[m$k$NN]{
   \includegraphics[width=0.43\textwidth]{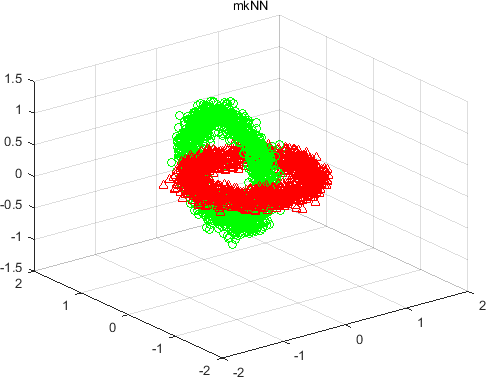}}
    \caption{Experimental results on synthetic data sets. (a)-(d): results of $k$NN, w$k$NN, g$k$NN and m$k$NN on data set 3.}
    \label{Resultknot}
\end{figure}

We first conduct experiments on three synthetic data sets shown in Figure \ref{SynDataSet} to demonstrate the superiority of  $mk\text{NN}$ over other $k\text{NN}$ algorithms. For each data set in Figure \ref{SynDataSet}, the three red triangles and the three green dots are the labeled samples of the two classes, respectively. Note that all  three data sets contain some ambiguous points (or bridging points) in the gap between two classes, making the classification even more challenging. Experimental results on these data sets are shown in Figures \ref{ResultTwoCircle} to \ref{Resultknot}.

From these results, we can see that because $k$NN uses Euclidean distance to determine the class label and Euclidean distance is not a proper similarity measure on the manifold, the results given by traditional $k$NN are quite erroneous. By introducing a weight scheme, w$k$NN can perform better than $k$NN, but the improvement is still quite limited. g$k$NN has much better results because geodesic distance on the manifold is a valid similarity measure. However, as mentioned in Section 1, graph distance tends to be sensitive to short-circuit phenomenon (i.e. the noise points lying between the two classes), thus g$k$NN still misclassifies many samples due to the existence of noisy points. In contrast, m$k$NN achieves the best results, because TRW takes all the possible graph paths into consideration, thus it embodies the global geometrical structure information of the manifold and can be much more effective and robust to noisy points.

\begin{figure}[H]
   \subfigure[Dataset 1]{
   \includegraphics[width=0.31\textwidth]{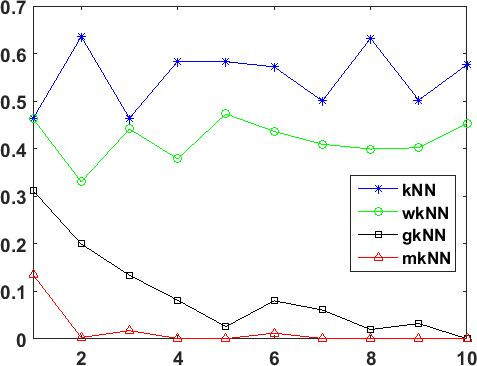}}
   \subfigure[Dataset 2]{
   \includegraphics[width=0.31\textwidth]{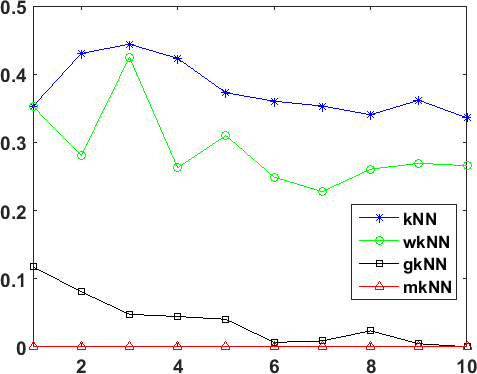}}
   \subfigure[Dataset 3]{
   \includegraphics[width=0.31\textwidth]{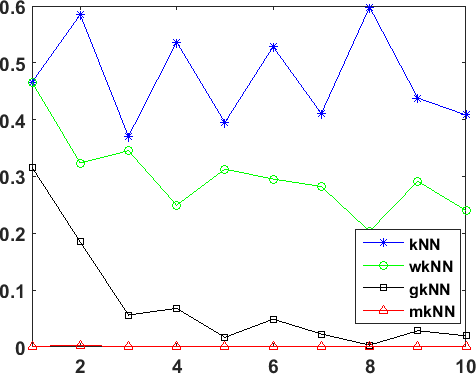}}
  \caption{Experimental results on synthetic data sets. Mean error rate on the ordinate and $k$ on the abscissa.}
  \label{SynDatasets_MeanErr}
\end{figure}

Figure \ref{SynDatasets_MeanErr} plots the mean error rate of each algorithm over 10 runs  on these data sets, as $k$ changes from 1 to 10. In each run, $k$ labeled samples are randomly selected from each class and the rest of the samples in the data set are treated as unlabeled samples to form the testing set.
From the results in Figure \ref{SynDatasets_MeanErr} we can see that m$k$NN performs significantly better than other version of $k$NN algorithms. It is interesting to note that on these manifold distributed data sets, while the mean error rates of $k$NN and w$k$NN have no obvious reduction as $k$ increases, the mean error rates of g$k$NN and m$k$NN decrease quickly. This indicates that g$k$NN and m$k$NN are able to exploit the information contained in the labeled samples in a more “efficient” way, because they take the manifold structure information into account. Again, m$k$NN achieves the lowest error rate.

In order to demonstrate the effectiveness of the weight reconstruction method, we run the algorithm on these data sets to reconstruct each sample and its TRW weight using its $k$ nearest neighbours with equation (1.7) and (1.8), respectively. The relative mean square error (RMSE) of the reconstruction result is computed by
$$RMSE(T,R)=\frac{{{\left\| T-R \right\|}^{2}}}{{{\left\| T \right\|}^{2}}}\times 100\%$$
where $T$ is the ground truth and $R$ is the reconstructed result. The results are shown in Table \ref{ReconstructRMSE}.
\begin{table}[h]
\centering  
\footnotesize
\caption{Reconstruction error of the synthetic data sets (\%).}
\label{ReconstructRMSE}
\begin{tabular}{ c c c c c} \toprule \hline
                                                                                & data set 1      & data set 2      &data set 3     &     \\ \cmidrule(r){2-4}
\emph{RMSE($X$, $\hat X$)}                                  & 0.0724             & 0.2734              &0.1606       &\\ \cmidrule(r){2-4}
\emph{RMSE($P_{TRW}$, $\hat P_{TRW}$)}      & 0.5075               & 1.0524                &1.9061          & \\ \hline \bottomrule
\end{tabular}

\end{table}
From these results we can see that the reconstruction method in equation (\ref{OptWeight}) and (\ref{ReconstructWeight}) can produce a very accurate approximation to the true samples and weights, respectively, on nonlinear manifold data sets.

\subsection{Experimental results on real-world data sets}
We also conduct experiments on six real-world data sets from the UCI data repository, which contains real application data collected in various fields and is widely used to test  the performance of different machine learning algorithms\footnote{UC Irvine Machine Learning Repository: http://archive.ics.uci.edu/ml/.}.  The information on these six data sets is listed in Table \ref{DataInfo} ($n$: number of samples; $d$: feature dimension; $C$: class number).

\begin{figure}[H]
  \subfigure[usps]{
   \includegraphics[width=0.47\textwidth]{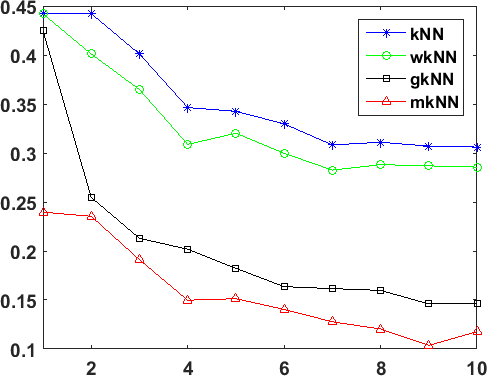}}
   \subfigure[segmentation]{
   \includegraphics[width=0.47\textwidth]{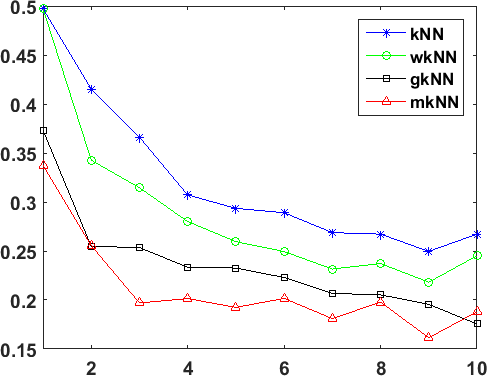}}

   \subfigure[banknote]{
   \includegraphics[width=0.47\textwidth]{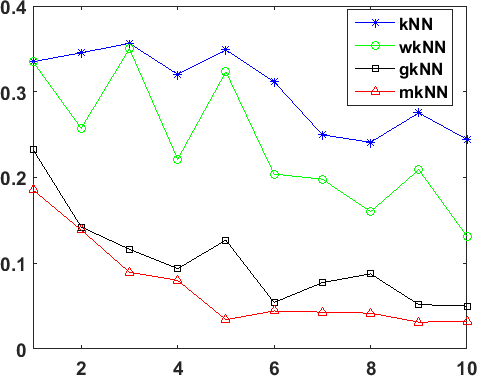}}
   \subfigure[pendigits]{
   \includegraphics[width=0.47\textwidth]{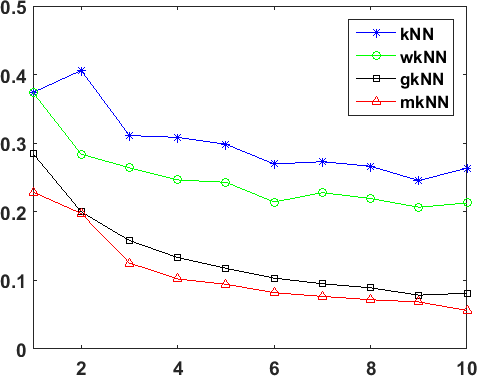}}

   \subfigure[multifeature]{
   \includegraphics[width=0.47\textwidth]{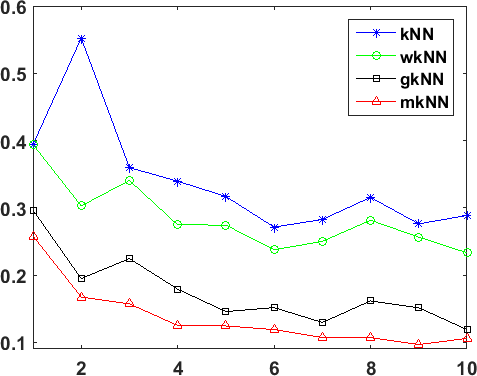}}
   \subfigure[satlog]{
   \includegraphics[width=0.47\textwidth]{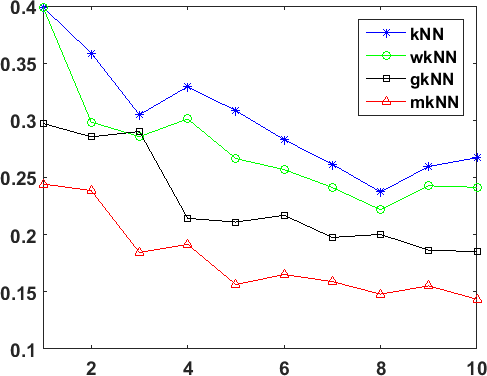}}
  \caption{Experimental results on six real-world data sets. Mean error rate on the ordinate and $k$ on the abscissa.}
  \label{RealWorldResults}
\end{figure}

\begin{table}[h]
\centering  
\footnotesize
\caption{Information on the experimental data sets.}
\label{DataInfo}
\begin{tabular}{ c c c c c c c} \toprule \hline
      & usps      & segmentation      &banknote      &pendigits     & multifeature         &statlog    \\ \cmidrule(r){2-7}
\emph{n}   & 9298      & 2086      &1348      &10992     & 2000         &6435    \\ \cmidrule(r){2-7}
\emph{d}   & 256      & 19   &4  &16     & 649           & 36  \\ \cmidrule(r){2-7}
\emph{C}   & 10      & 6   &2  &10     &10            &6  \\ \hline \bottomrule
\end{tabular}

\end{table}
On each data set, we let $k$ change from 1 to 10. For each $k$, we run each algorithm 10 times, with a training set containing $k$ labeled samples randomly selected from each class and a testing set containing all the rest samples. The final error rate is the mean of the 10 error rates and is shown in Figure \ref{RealWorldResults}.

From Figure \ref{RealWorldResults}, we can see that the proposed m$k$NN outperforms the other algorithms in terms of both  the accuracy and stability of  performance. The error curves of m$k$NN decrease much more quickly than that the other algorithms, which indicates that m$k$NN is able to utilize fewer labeled samples to achieve better accuracy. This is very important in applications where there is a very small amount of labeled samples available, because labeled samples are usually more expensive to obtain, i.e. they need to be annotated by human with expert knowledge and/or special equipment and take a lot of time.  Therefore, it is of great practical value that a classifier is capable of accurately classifying abundant unlabeled samples, given only a small number of labeled samples.

From Figure \ref{RealWorldResults}, we can also conclude that: (1) by introducing a weight scheme in the traditional $k$NN algorithm,  weighted $k$NN (w$k$NN) can generally outperform traditional $k$NN; (2) the performance of geodesic $k$NN (g$k$NN) has a relative large improvement to both $k$NN and weighted $k$NN for most cases, because geodesic distance is a valid measure on manifold; (3) the manifold $k$NN (m$k$NN) almost always achieves the smallest error rate, because the constrained TRW is a more effective and robust measure of the global geometrical structure on manifold and, meanwhile, m$k$NN can take both the class information and constraint information into account.
\subsection{Experimental results of the comparison with other traditional supervised classifiers}
 We  conduct experiments to compare the performance of the proposed manifold $k$NN algorithm with other popular supervised classifiers on the six real-world data sets. The baseline algorithms are: Support Vector Machine (SVM)\footnote{For SVM we use the libsvm toolbox: \url{https://www.csie.ntu.edu.tw/~cjlin/libsvm/}. Other algorithms we use MATAB toolbox.}, Artificial Neural Networks (ANN), Naive Bayes (NB) and Decision Tree (DT). The configurations of the baseline algorithms are: radial basis kernel for SVM; three-layer  networks trained with back-propagation for ANN; kernel smoothing density estimation for NB; binary classification tree and merging-pruning strategy according to validation error for DT. We adopt a grid-search strategy to tune the parameters of each algorithm and the parameters are set to  produce the lowest 2-fold cross validation error rate.

 To examine the capability of classifying plenty of unlabeled samples with only a few labeled samples, we randomly choose three  labeled samples from each class to form the training set and the remaining samples are treated as unlabeled samples to form the testing set. Each algorithm runs 10 times and the final result is the average of these 10 error rates.  The experimental results (mean error $\pm$ standard deviation) are shown in Table \ref{TabResultRealXU}.
\begin{table}[h]
\caption{Experimental results on real-world data sets (error rate: \%)}
\label{TabResultRealXU}
\footnotesize
\centering  
\settowidth \rotheadsize{\theadfont Second multilined}
\begin{tabular}{l c c c c c}\\
\toprule \hline
{} &SVM &ANN &NB & DT& m$k$NN \\ \cmidrule(r){2-6} 
usps &77.35$\pm$4.6 & 40.96$\pm$4.3 & 34.73$\pm$4.9 & 73.96$\pm$8.4 & \textbf{19.37$\pm$3.9} \\ \cmidrule(r){2-6}
segmentation &66.54$\pm$8.7 &42.54$\pm$3.8 &50.83$\pm$4.3&49.55$\pm$2.6&\textbf{24.18$\pm$4.5} \\ \cmidrule(r){2-6}
banknote &70.01$\pm$14.7 &23.03$\pm$6.2 &28.37$\pm$7.8&44.43$\pm$0.4&\textbf{9.73$\pm$6.9}\\ \cmidrule(r){2-6}
pendigits &49.41$\pm$12.3 &44.02$\pm$27.4 &42.36$\pm$6.2&80.83$\pm$11.3 &\textbf{12.51$\pm$2.3}\\ \cmidrule(r){2-6}
multifeature &82.89$\pm$6.4 &50.53$\pm$24.2 &26.89$\pm$2.9&76.14$\pm$2.9&\textbf{15.36$\pm$2.3} \\ \cmidrule(r){2-6}
satlog &57.26$\pm$17.1 &30.15$\pm$4.9 &28.81$\pm$5.9&72.55$\pm$13.5&\textbf{21.07$\pm$8.0} \\
\hline \bottomrule\\
\end{tabular}
\end{table}

From these results, we can see that m$k$NN  significantly outperform  these traditional supervised classifiers, given only three labeled samples per class. Similar to traditional $k$NN, traditional supervised classifiers are incapable of exploiting the manifold structure information of the underlying data distribution, thus their accuracies are very low while the labeled sample number is very small. Furthermore, when the labeled samples are randomly selected, their positions vary greatly in data space. Sometime they are not uniformly distributed and thus cannot well cover the whole data distribution. As a result, the performance of traditional classifiers also varies greatly. In contrast, because m$k$NN adopts tired random walk to measure manifold similarity which reflects the global geometrical  information of the underlying manifold structure and is robust to local changes. \cite{tu2014novel}, it can achieve much better results.

\begin{figure}[H]
  \subfigure[usps]{
   \includegraphics[width=0.47\textwidth]{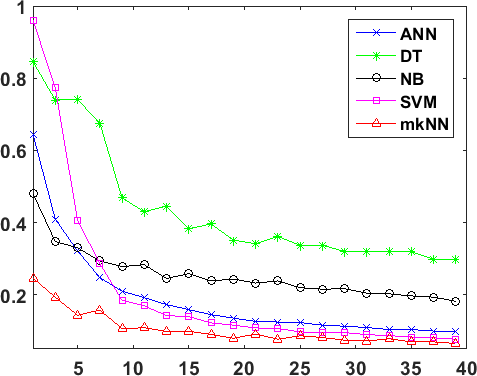}}
   \subfigure[segmentation]{
   \includegraphics[width=0.47\textwidth]{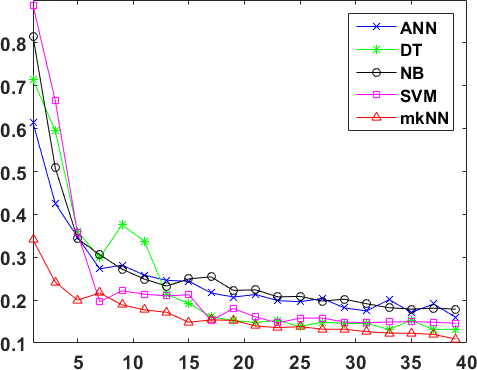}}

   \subfigure[banknote]{
   \includegraphics[width=0.47\textwidth]{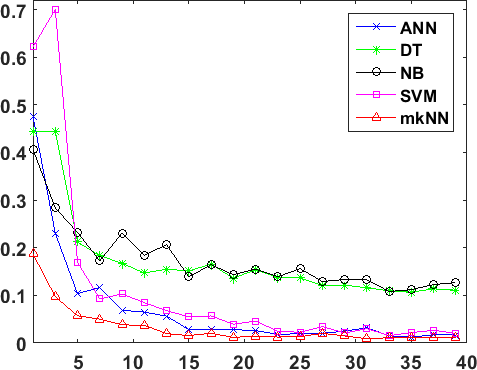}}
   \subfigure[pendigits]{
   \includegraphics[width=0.47\textwidth]{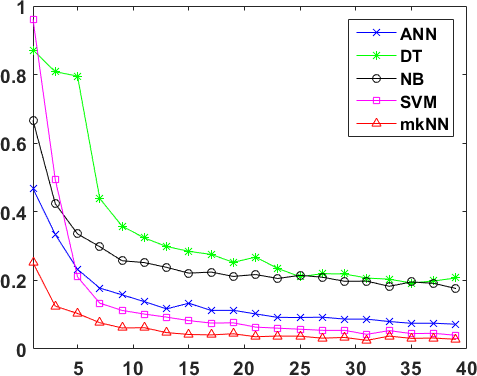}}

   \subfigure[multifeature]{
   \includegraphics[width=0.47\textwidth]{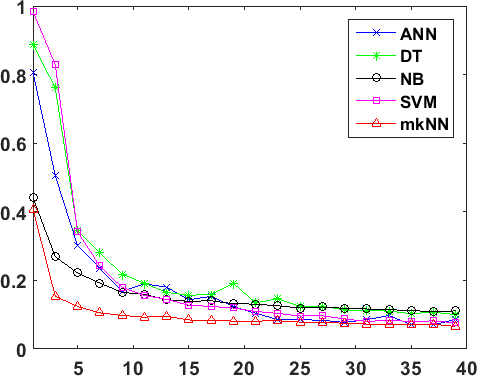}}
   \subfigure[satlog]{
   \includegraphics[width=0.47\textwidth]{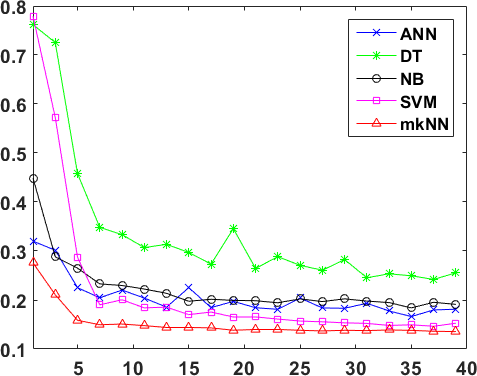}}
  \caption{Experimental results on six real-world data sets. Mean error rate on the ordinate and  labeled samples number per class  on the abscissa.}\label{RealWorldResults_mean}
\end{figure}

\begin{figure}[H]
  \subfigure[usps]{
   \includegraphics[width=0.47\textwidth]{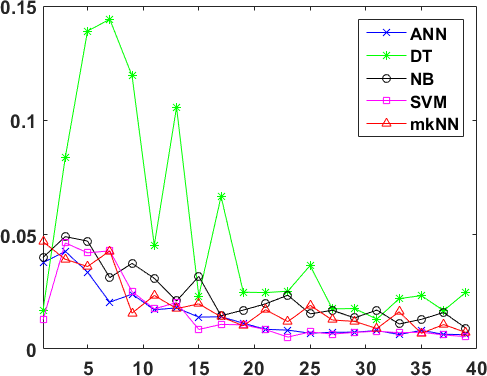}}
   \subfigure[segmentation]{
   \includegraphics[width=0.47\textwidth]{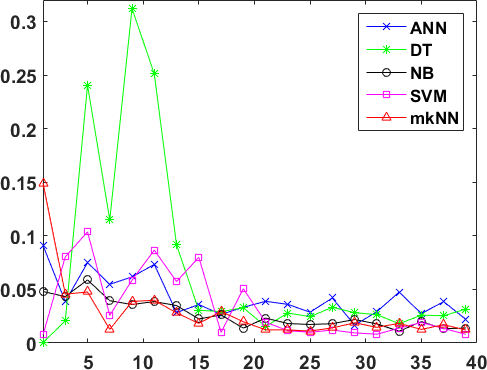}}

   \subfigure[banknote]{
   \includegraphics[width=0.47\textwidth]{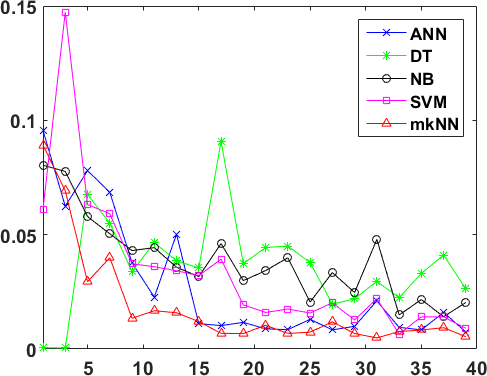}}
   \subfigure[pendigits]{
   \includegraphics[width=0.47\textwidth]{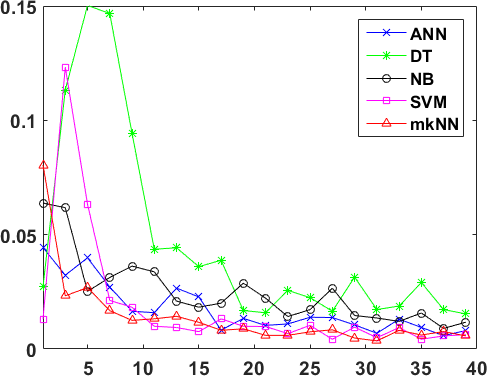}}

   \subfigure[multifeature]{
   \includegraphics[width=0.47\textwidth]{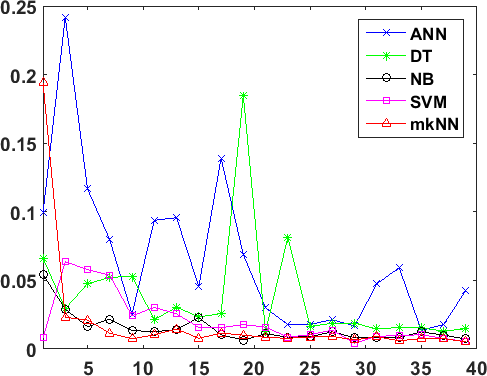}}
   \subfigure[satlog]{
   \includegraphics[width=0.47\textwidth]{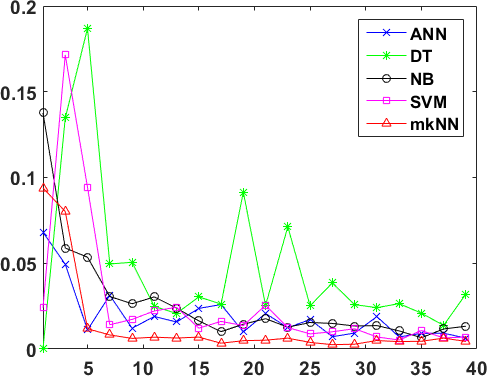}}
  \caption{Experimental results on six real-world data sets. Standard deviation of error rates on the ordinate and labeled samples number per  class on the abscissa.}\label{RealWorldResults_std}
\end{figure}

To further investigate the performance of these algorithms under the condition of providing different number of training samples, we carry out experiments with different training set sizes. For each data set, we let $k$ varies from 1 to 40 ($k=1, 3, ..., 39$) and randomly choose $k$ labeled samples from each class to form the training set. The rest of the data set are treated as unlabeled samples to test each algorithm's performance. For each $k$, every algorithm runs 10 times on each data set.  The mean error rate and standard deviation of the 10-run results are shown in Figures \ref{RealWorldResults_mean} and \ref{RealWorldResults_std}, respectively.

From Figure \ref{RealWorldResults_mean} we can see that while the labeled sample number is small, the error rates of the traditional algorithms are very large (as also indicated in Table \ref{TabResultRealXU} which contains the results of the three labeled samples per class). As the number of labeled sample increases, the error rates decrease. This trend becomes slower after 15 labeled samples per class. The proposed m$k$NN achieves the lowest error rate for both situations of small and large number of labeled samples. The improvement is especially obvious and significant when the labeled samples number is small. From Figure \ref{RealWorldResults_std}, we can also see that the performance of m$k$NN is also relatively steady for most cases. One should note that on the left of each plot in Figure \ref{RealWorldResults_std}, the small standard deviation values of decision tree (DT) and SVM are due to the fact that their error rates are always very high in each run (e.g. the error rates of SVM on data set \emph{usps} concentrate closely around 95).

\subsection{Experimental results of time complexity}
\begin{figure}[H]
  \subfigure[Satlog data set]{
   \includegraphics[width=0.45\textwidth]{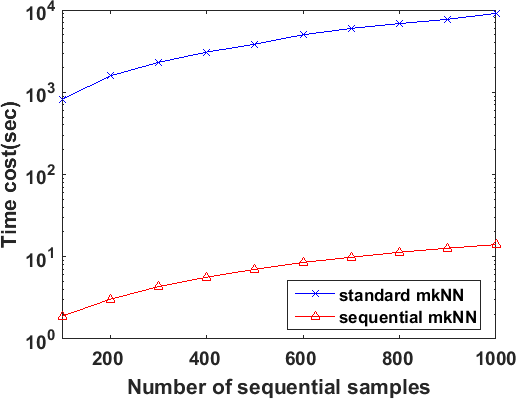}
   \includegraphics[width=0.45\textwidth]{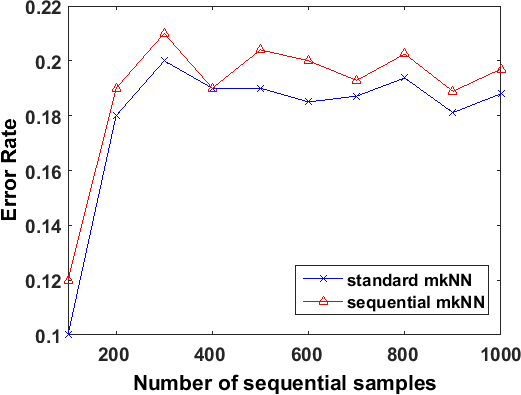}}

   \subfigure[Banknote data set]{
      \includegraphics[width=0.45\textwidth]{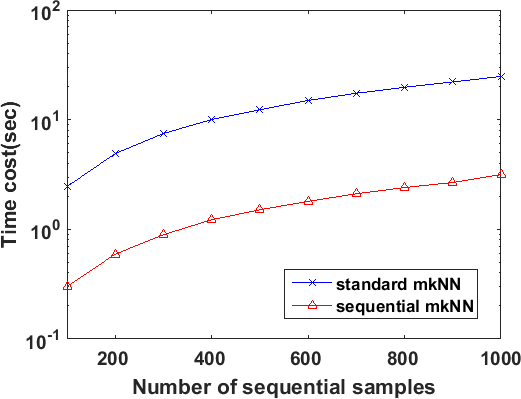}
   \includegraphics[width=0.45\textwidth]{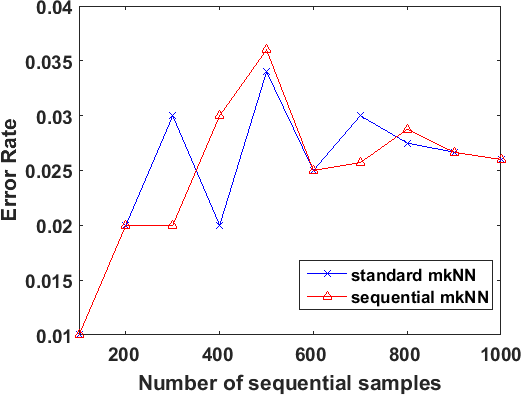}}

   \subfigure[Pendigits data set]{
      \includegraphics[width=0.45\textwidth]{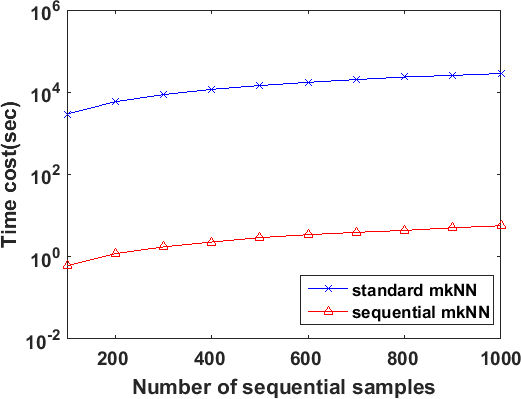}
   \includegraphics[width=0.45\textwidth]{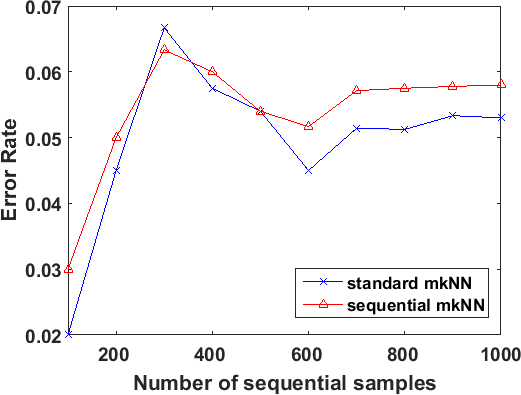}}
    \caption{Experimental results of the comparison between standard m$k$NN and sequential m$k$NN to classify online samples real-world data sets.}\label{SeqExperiment}
\end{figure}

To examine the effectiveness and efficiency of the proposed sequential learning strategy in Section 5, we conduct experiments on  three real-world data sets \emph{banknote,satlog} and $pendigits$ to show the difference of m$k$NN's performance with and without sequential learning algorithm. Each data set is divided into three subsets: training set, validation set and online set. The training set is fixed to contain 10 labeled samples per class. We conduct 10 experiments, with the online set size changes from 100 to 1000 with step size 100. The validation set contains the rest of the samples. First, m$k$NN  runs on training set and validation set to learn the class distribution. Thereafter, at each time a sample is drawn from the online set as a new coming sample. For sequential m$k$NN, the previous learning result is used  to classify the new sample according to  equation (\ref{OptWeight}) and equation (\ref{ReconstructWeight}). For standard m$k$NN, the new sample is added to the validation set and the whole learning process is repeated to classify the new sample. The experimental results\footnote{The configuration of our computer: 16GB RAM, double-core 3.7GHz Intel Xeon CPU and MATLAB 2015 academic version.} are shown in Figure \ref{SeqExperiment}.

From these results, we can see that while the classification accuracy of standard m$k$NN and sequential m$k$NN are comparable, the time cost of sequential m$k$NN reduces dramatically for online classification (e.g., for \emph{satlog} data set, to classify 1000 sequentially coming samples, standard m$k$NN takes about 9100 seconds but sequential m$k$NN uses only about 14 seconds to achieve a similar result). Therefore, the sequential algorithm has great merit in solving the online classification problem and can be potentially applied to a wide range of transductive learning algorithms to make them inductive.

We also conduct experiments to compare the time complexity of all the baseline algorithms with the proposed m$k$NN on these three data sets. Each data set is split  into \emph{training} and \emph{testing} parts three times independently, with splitting ratios 10\%, 25\% and 50\%, respectively. For each split, every  algorithm runs 10 times and the mean time cost is recorded. Table \ref{TimeCostExperiment} shows the experimental results (the second column shows the splitting ratio).

\begin{table}[ht]
\caption{Comparison of overall time for training and testing on three data sets (sec)}
\label{TimeCostExperiment}
\centering  
\renewcommand\arraystretch{0.4} 
\footnotesize
\begin{tabular*}{0.93\textwidth}{@{}ccrrrrrrrr}\toprule \midrule
data     &  (\%)     & SVM     &DT     &ANN        &NB   &$k$NN & w$k$NN & g$k$NN &m$k$NN\\ \midrule \midrule
           &10          &0.01     &0.14    & 0.23      &0.09       &0.03          &0.02        &1.23      &0.19    \\ \cmidrule(r){3-10}
banknote&25       &0.01     &0.19    & 0.24      &0.09        &0.04         &0.03        &1.23    &0.20    \\  \cmidrule(r){3-10}
           &50          &0.02     &0.29    & 0.27      &0.08        &0.05        &0.05         &1.26     &0.20   \\ \midrule
           &10          &0.47     & 0.43    & 0.59      &7.14       &0.13         &0.14        &31.99    &6.01    \\ \cmidrule(r){3-10}
satlog  &25          &1.78     &1.16    & 1.58      &7.56         &0.24         &0.29      &32.55    &6.08    \\  \cmidrule(r){3-10}
           &50         &5.12     &3.11    & 4.01     &7.61           &0.31         &0.40      &32.82     &6.30    \\ \midrule
           &10         &0.44     &0.78    & 0.70      &8.70          &0.22         &0.28      &149.14   &25.27   \\ \cmidrule(r){3-10}
pendigits&25         &0.86     &2.28    & 2.65      &8.46        &0.41         &0.59      &146.89   &26.11    \\  \cmidrule(r){3-10}
           &50         &1.55     &6.09   & 5.06      &6.93           &0.54         &0.80       &145.71    &25.35    \\ \midrule  \bottomrule
\end{tabular*}
\end{table}

From this table we can see that the algorithms fall into two categories according to their time costs: one category contains SVM, DT, ANN, $k$NN and w$k$NN, whose time costs are closely related with the training data set size. Another category includes NB, g$k$NN and m$k$NN, whose time costs depend more upon the overall data set size.   SVM, $k$NN and w$k$NN are generally faster than others. Although m$k$NN is the second slowest one,  it is still much faster than g$k$NN and its overall time does not prolong as the training set size increases.  It should be mentioned that although the classification accuracy of m$k$NN is much better than  $k$NN and other traditional supervised classifiers,  the computational complexity of m$k$NN is also higher ($O(n^3)$) than traditional $k$NN ($O(n)$). Directly computing   $P_{TRW}$ matrix is time cost, since it is not   symmetrical, positive definite matrix and thus a LU decomposition has to be used. One way to speed-up matrix inverse is to convert  $P_{TRW}$ to a symmetrical, positive definite matrix and then adopt the Cholesky decomposition, whose computational cost is just a half of the LU decomposition \cite{golub2012matrix}. Noting that  ${{P}_{TRW}}={{\left( I-\alpha {{D}^{-1}}W \right)}^{-1}}={{D}^{1/2}}{{\left( I-\alpha {{D}^{-1/2}}W{{D}^{-1/2}} \right)}^{-1}}{{D}^{-1/2}}={{D}^{1/2}}{{R}^{-1}}{{D}^{-1/2}}$, where $R=I-\alpha {{D}^{-1/2}}W{{D}^{-1/2}}$ is  a symmetrical, positive definite matrix\footnote{$R$ is apparently symmetrical because $W$ is symmetrical (Let $W=(W+W^T)/2$ if $W$ is not symmetrical). We prove its positive definiteness. Because the spectral radius of $P_{TRW}$ is in (0, 1) and $P_{TRW}$ is similar to $R^{-1}$, so spectral radius of $R^{-1}$ is in (0, 1). So all eigenvalues of $R$ are positive. Therefore $R$ is a positive definite, symmetrical matrix.}, we can first  inverse matrix $R$ using Cholesky decomposition and then compute $P_{TRW}$ with very small computational cost. Our following work will be focused on  the further reduction of computational complexity.
\section{Conclusions}
In this paper we proposed a new $k$ nearest-neighbor algorithm, m$k$NN, to classify nonlinear manifold distributed data as well as traditional Gaussian distributed data, given a very small amount of labeled samples. We also presented an algorithm to attack the problem of high computational cost for classifying  online data with m$k$NN and other transductive algorithms. The superiority of the m$k$NN has been demonstrated by substantial experiments on both synthetic data sets and real-world data sets.  Given the widespread appearance of manifold structures in real-world problems and the popularity of the traditional $k$NN algorithm, the proposed manifold version $k$NN shows promising potential for classifying manifold-distributed data.

\textbf{Acknowledgements}: The authors wish to thank the anonymous reviewers for reading the entire manuscript and offering many useful suggestions. This research is partly supported by NSFC, China (No: 61572315) and 973 Plan，China (No. 2015CB856004). The research is also partly supported by YangFan Project (Grant No. 14YF1411000) of Shanghai Municipal Science and Technology Commission, the Innovation Program (Grant No. 14YZ131) and the Excellent Youth Scholars (Grant No. sdl15101) of Shanghai Municipal Education Commission, the Science Research Foundation of Shanghai University of Electric Power (Grant No. K2014-032).



\bibliographystyle{elsarticle-harv}
\bibliography{refs}






\end{document}